\documentclass[10pt,twocolumn,letterpaper]{article}

\usepackage{cvpr}            
\usepackage{graphicx}
\usepackage{amsmath}
\usepackage{amssymb}
\usepackage{booktabs}
\usepackage{bm}
\usepackage{multirow}
\usepackage{makecell}

\usepackage[pagebackref,breaklinks,colorlinks]{hyperref}
\usepackage[capitalize]{cleveref}
\crefname{section}{Sec.}{Secs.}
\Crefname{section}{Section}{Sections}
\Crefname{table}{Table}{Tables}
\crefname{table}{Tab.}{Tabs.}

\begin{document}

\title{Self-Supervised Vision Transformer for Enhanced Virtual Clothes Try-On}

\author{
\textnormal{Lingxiao Lu}$^1$, 
\textnormal{Shengyi Wu}$^1$,
\textnormal{Haoxuan Sun}$^1$, 
\textnormal{Junhong Gou}$^1$, \\ 
\textnormal{Jianlou Si}$^2$,
\textnormal{Chen Qian}$^2$, \textnormal{Jianfu Zhang}$^1$\thanks{Corresponding authors.}~, \textnormal{Liqing Zhang}$^{1*}$\\
$^1$Shanghai Jiao Tong University, $^2$SenseTime\\
\tt \small \{lulingxiao,wsykk2,guwangtu,goujunhong,c.sis,zhang-lq\}@sjtu.edu.cn \\ 
\tt \small \{sijianlou,qianchen\}@sensetime.com
}

\maketitle

\begin{abstract}
Virtual clothes try-on has emerged as a vital feature in online shopping, offering consumers a critical tool to visualize how clothing fits. In our research, we introduce an innovative approach for virtual clothes try-on, utilizing a self-supervised Vision Transformer (ViT) coupled with a diffusion model. Our method emphasizes detail enhancement by contrasting local clothing image embeddings, generated by ViT, with their global counterparts. Techniques such as conditional guidance and focus on key regions have been integrated into our approach. These combined strategies empower the diffusion model to reproduce clothing details with increased clarity and realism. The experimental results showcase substantial advancements in the realism and precision of details in virtual try-on experiences, significantly surpassing the capabilities of existing technologies.
\end{abstract}

\section{Introduction}
The integration of virtual clothes try-on functionality in online shopping platforms has become a key feature, providing consumers with an invaluable tool to visualize and evaluate the fit of clothing items prior to purchase.
With the growing trend of online shopping, the advancement of sophisticated virtual try-on techniques is crucial in elevating the overall user experience \cite{Wang_Zheng_Liang_Chen_Lin_Yang_2018,Cai_Xiong_Xu_Wang_Li_Pan_2022}. 
These developments play a significant role in bridging the gap between physical and online retail experiences, catering to the evolving needs of consumers and reshaping the landscape of the retail industry.

A considerable portion of previous research in virtual try-on has heavily leaned on Generative Adversarial Networks (GANs) to generate lifelike images \cite{Cai_Xiong_Xu_Wang_Li_Pan_2022}. To refine the preservation of intricate features, earlier studies \cite{Minar_Tuan_Ahn_Rosin_Lai,Wang_Zheng_Liang_Chen_Lin_Yang_2018,Han_Wu_Wu_Yu_Davis_2018,Yang_Zhang_Guo_Liu_Zuo_Luo_2020,Ge_Song_Zhang_Ge_Liu_Luo_2021} have integrated specialized warping components. 
These components are tailored to align target clothing accurately with human figures. 
Subsequently, the warped clothing, combined with a representation of the individual agnostic to their garments, is input into GAN-based generators to yield final visual outputs.  
Additionally, certain efforts \cite{Choi_Park_Lee_Choo_2021,Lee_Gu_Park_Choi_Choo_2022} have extended these methodologies to accommodate high-resolution imagery. It is essential to acknowledge, however, that the efficacy of such methods heavily hinges on the quality of the garment warping process.
Recently, diffusion models have emerged as notable alternatives to traditional generative models \cite{Ho_Jain_Abbeel_Berkeley,Rombach_Blattmann_Lorenz_Esser_Ommer_2022,Song_Meng_Ermon_2020,saharia2022palette}, recognized for their comprehensive distribution coverage, well-defined training objectives, and enhanced scalability \cite{Nichol_Dhariwal_Ramesh_Shyam_Mishkin_McGrew_Sutskever_Chen}. The Stable Diffusion Network, in particular, has gained prominence for its ability to create lifelike images by leveraging the reverse diffusion process \cite{Lee_Gu_Park_Choi_Choo_2022}. 
However, a significant challenge in current approaches is the limited capacity to accurately and authentically replicate complex clothing details. 
This limitation is highlighted in the constraints of struggling to maintain precise control over crucial details that are vital for a realistic virtual try-on experience. The is because that previous methods \cite{morelli2023ladi,Yang_Gu_Zhang_Zhang_Chen_Sun_Chen_Wen_2022}
employ CLIP \cite{Radford_Kim_Hallacy_Ramesh_Goh_Agarwal_Sastry_Amanda_Mishkin_Clark_et} to extract detailed information from the reference image as guidance for Stable Diffusion (SD). 
However, CLIP's inclination towards semantic description often results in pronounced differences between dissimilar semantic descriptions but limited variation within identical text prompts. Consequently, this leads to reduced distinctiveness for similar semantic meanings, posing a significant challenge in virtual try-on scenarios where most clothing items possess similar textual descriptions, making it challenging to accurately describe specific designs and layouts.

In our proposed methodology, we overcome the limitations of current virtual clothes try-on techniques by introducing innovative strategies focused on improving detail accuracy and realism.
To overcome the constraints of textual descriptions, it becomes crucial to extract effective information from clothing datasets to create optimal conditions for SD to produce superior images.
Inspired by previous self-supervised transformer works \cite{Caron_Touvron_Misra_Jegou_Mairal_Bojanowski_Joulin_2021}, which utilizes self-supervised learning for visual representation, our approach follows suit. 
The training methodology revolves around contrasting representations from different perspectives of the same clothes, facilitated by a dual training framework comprising both teacher and student networks. 
This method enables the extraction of semantically rich features from clothing data, significantly enhancing the condition quality for SD. 
Furthermore, in applying a self-supervised approach to Vision Transformers (ViT) \cite{Dosovitskiy}, we discovered that a pre-trained ViT exhibits limited adaptability to our dataset and lacks the capacity to express clothing-specific features adequately. 
To address this, we intentionally fine-tune ViT on our dataset, concentrating on essential clothing aspects such as collars, sleeves, text, and patterns. By utilizing the self-attention maps of ViT to identify and concentrate on key clothing elements, we selectively sample local crops that surround these crucial keypoints. 
This approach enriches the network with detailed information necessary for an accurate representation of clothing, ensuring that ViT precisely identifies and emphasizes critical details during the fine-tuning process.
Importantly, our method introduces only minor additional inference time compared to other methods based on diffusion models.
Through the introduction of these innovative methodologies and enhancements, we aim for our proposed approach to deliver substantial improvements in the realism and detail precision of virtual try-on experiences, positioning it as a superior alternative to current solutions.

\section{Related Works}

\subsection{Virtual Try-On }
Virtual try-on technology has increasingly become a focal point in research, driven by its potential to revolutionize the consumer shopping experience. The domain encompasses two primary methodologies: 2D and 3D approaches, as outlined in previous studies \cite{He_Song_Xiang_2022}. The 3D virtual try-on \cite{chongsig21,zhao2021m3d,korosteleva2023garmentcode,liu2023towards}, offering an immersive user experience, requires sophisticated 3D parametric human models and extensive 3D datasets, often entailing significant costs. Conversely, 2D virtual try-on methods are more accessible and prevalent in practical applications due to their lighter computational demands. Prior research in 2D virtual try-on \cite{Han_Wu_Wu_Yu_Davis_2018,Minar_Tuan_Ahn_Rosin_Lai,Wang_Zheng_Liang_Chen_Lin_Yang_2018}, has concentrated on adapting clothing to fit human figures flexibly, yet it faces challenges in managing substantial geometric deformations. Furthermore, flow-based methods \cite{Han_Huang_Hu_Scott_2019,Ge_Song_Zhang_Ge_Liu_Luo_2021,He_Song_Xiang_2022} have been explored to depict the appearance flow field between clothing and the human body, aiming to enhance the fit of garments. While these methods offer significant advancements, they frequently fall short in generating high-quality, high-resolution imagery, particularly when it comes to replicating intricate details of clothing.

Moreover, with increasing resolution, synthesis stages based on Generative Adversarial Networks (GANs) \cite{Lee_Gu_Park_Choi_Choo_2022,Choi_Park_Lee_Choo_2021,lin2023fashiontex,lewis2021tryongan} encounter difficulties in preserving the integrity and characteristics of clothing, resulting in reduced realism and an escalation of artifacts. The generative capabilities of GANs, particularly in conjunction with high-resolution clothing warping, are constrained, impacting the overall quality of the synthesized images. These challenges underscore the need for more sophisticated approaches capable of surpassing the limitations inherent in current virtual try-on methods. Employing diffusion models emerges as a promising solution, offering significant improvements in virtual try-on performance. Specifically, diffusion models excel in managing complex clothing attributes and achieving realistic image synthesis, addressing the critical shortcomings of existing technologies.

\subsection{Virtual Try-On Methods Based on Diffusion Models}
In recent years, Diffusion Models \cite{Ho_Jain_Abbeel_Berkeley,Song_Meng_Ermon_2020} have been proposed to generate realistic images by reversing a gradual noising process. And latent diffusion models \cite{Rombach_Blattmann_Lorenz_Esser_Ommer_2022}, trained on latent space representations, incorporate cross-attention layers within their network structure to handle generic conditions as input. This advancement has established diffusion models as formidable rivals to GANs in image synthesis tasks. Recent studies have also delved into using text information as a condition in the denoising process, enabling diffusion models to create images with text-relevant features \cite{Ramesh_Dhariwal_Nichol_Chu_Chen,voynov2023sketch,guerrero2024texsliders,li2023layerdiffusion,chefer2023attend}. 

LaDI-VTON \cite{morelli2023ladi} introduces a novel textual inversion component, mapping the visual attributes of in-shop garments to CLIP token embeddings to generate pseudo-word token embeddings, conditioning the generation process and preserving detailed textures. Furthermore, the development of spatial-level guidance techniques in diffusion models \cite{Meng_Song_Song_Wu_Zhu_Ermon_2021,Choi_Kim_Jeong_Gwon_Yoon_2021}, particularly through targeted interventions in the denoising phase, has significantly enhanced their applicability across various domains.
Despite these advancements, effective implementation of virtual try-on using diffusion models remains a challenge. Text-to-image methods \cite{Ramesh_Dhariwal_Nichol_Chu_Chen,Saharia_Chan_Saxena_Li_Whang_Denton_Kamyar_Ghasemipour_Karagol_Mahdavi_et} alone are insufficient for accurately depicting the varied appearances of clothing. Similarly, inpainting techniques \cite{Yang_Gu_Zhang_Zhang_Chen_Sun_Chen_Wen_2022} struggle with maintaining precise control over finer details.

In response to these limitations, our approach introduces a novel self-supervised component specifically designed to enhance detail accuracy and realism in virtual clothing try-ons, focusing on the extraction of critical keypoints.

\section{Method}
In this section, we will detail the framework of our method, covering the foundational aspects of the diffusion model. We will also introduce our Vision Transformer training process and the procedure for extracting key points of clothing.

\subsection{Overall Structure}

\begin{figure}[t]
\centering
\includegraphics[width=1\linewidth]{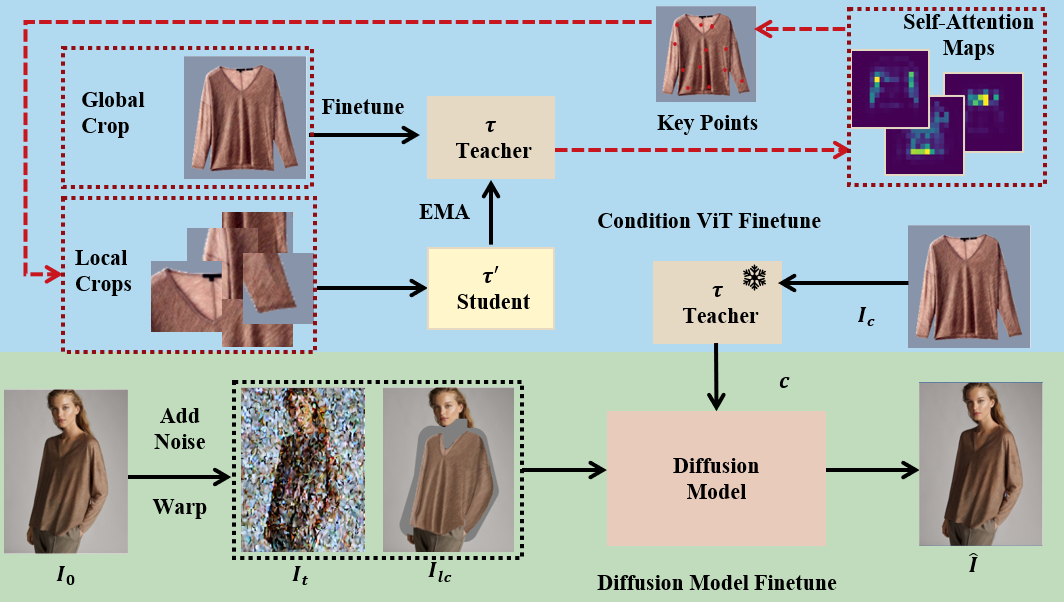}
\caption{Overall framework of our network. we utilize the Stable Diffusion (SD) Inpainting network and employ a specially finetuned Vision Transformer (ViT) to direct the network's focus towards intricate clothes image details. The finetuned ViT, denoted as $\tau$, also functions as an essential feature extractor, instrumental in calculating the loss and further refining the inpainting process. Alongside, we integrate warp features into the input to enhance the alignment between the network's internal features and those in the given condition. For simplicity in representation, we omit the encoder $E$ and the decoder $D$ of the SD network in our depiction.}
\label{fig:task}
\end{figure}

In this research, our objective is to harness diffusion models within an inpainting framework for virtual try-on tasks, focusing on intricacies of clothing such as sleeves, collars, and textual patterns. Previous methodologies have explored various approaches for injecting explicit information, yet they often overlook these critical clothing details. To address this, we introduce a self-supervised learning-based detail enhancer, designed to aid our network in better recognizing and integrating these essential features. 
As depicted in Figure~\ref{fig:task}, our methodology involves processing a person image $I_o \in \mathbb{R}^{H \times W \times 3}$ and a clothing image $I_c \in \mathbb{R}^{H' \times W' \times 3}$ to generate a synthesized image $\hat{I} \in \mathbb{R}^{H \times W \times 3}$. This image aims to realistically blend the person's attributes from $I_o$ with the clothing elements of $I_c$.
For try-on scenarios, we utilize a binary mask $m \in {0, 1}^{H \times W}$ to identify the areas needing inpainting, generally covering the upper body and arms of the person. In the resulting image $\hat{I}$, areas where $m = 0$ should faithfully reflect $I_p$, and those with $m = 1$ should seamlessly incorporate elements from $I_c$, ensuring a natural integration with the person's figure.

The clothing image $I_c$ plays a pivotal role in guiding the model to generate try-on images. It is input into a Vision Transformer (ViT) \cite{Dosovitskiy} to produce conditional embeddings $c$ that direct the diffusion models. To improve the accuracy of detail replication, we fine-tune ViT as a condition encoder $\tau$ using our specialized clothing dataset. 
The fine-tuning process adopts distillation approaches, involving the collaboration of teacher and student networks. Throughout this phase, the teacher model processes the entire global crop  $I_c$, whereas the student model processes both $I_c$ and its locally cropped segments. Such an arrangement allows the student model to effectively assimilate knowledge from the teacher, thereby promoting self-supervised learning within ViT. Post fine-tuning, the ViT teacher model, now adept at capturing details, is employed as a detail injector in our stable diffusion inpainting network. It inputs complete clothing images to ensure enhanced precision in the final output.

To facilitate improved interaction with ViT features, our model's input incorporates a warped image. During this warping process, the clothing image $I_c$ is input into a warping network, designed to predict an appearance flow field that aligns the clothing accurately. The outcome is a warped clothing image, which, when combined with a clothes-agnostic person image and a mask, yields a masked coarse result $I_{lc}$. 
The warping process in our framework adopts an iterative refinement strategy, adept at capturing long-range correspondences. This approach is particularly effective in addressing significant misalignments in clothing images. For a more comprehensive understanding of the warping techniques employed, please refer to \cite{Ge_Song_Zhang_Ge_Liu_Luo_2021,Han_Huang_Hu_Scott_2019}.

Subsequently, we will explore the intricacies of the diffusion model, our method for self-supervised ViT fine-tuning, and the content correction loss mechanism that is specifically developed based on ViT insights.

\subsection{Virtual Try-on by Diffusion Models}
Existing studies \cite{Yang_Gu_Zhang_Zhang_Chen_Sun_Chen_Wen_2022,morelli2023ladi} have demonstrated the considerable capabilities of diffusion models in try-on tasks, which is why our work also adopts a diffusion model, specifically leveraging the open-source StableDiffusion (SD) \cite{Lee_Gu_Park_Choi_Choo_2022} framework. The SD Network excels in generating realistic images by leveraging the reverse diffusion process. Starting with the target image $I_0$, an initial forward diffusion process denoted as $q(\cdot)$ is applied. This process incrementally introduces noise following a Markov chain, ultimately transforming the image into a Gaussian distribution. To optimize computational efficiency, a latent diffusion model is employed. This involves transforming images from the image space to the latent space using a pre-trained encoder $E$, followed by image reconstruction with a pre-trained decoder $D$. 
The forward diffusion process on the latent variable $z_0 = E(I_0)$ at any given timestamp $t$ is defined as:
\begin{equation}
    z_t = \sqrt{\alpha_t} z_0 + \sqrt{1 - \alpha_t} \epsilon,
\end{equation}
where
$\alpha := \prod_{s=1}^{t} (1-\beta_s)$ and $\epsilon \sim N(0, I)$.
Here, $\beta$ denotes a predefined variance schedule spanning $T$ steps. The latent code $z_{lc}$ is derived by passing $I_{lc}$ through the encoder $E$, which is then concatenated with the downsampled mask $m$ to form the input ${z_t, z_{lc}, m}$. For the denoising phase, an enhanced Diffusion UNet is utilized to predict a refined version of the input. Within this framework, the condition $c$ extracted from $I_c$, using a condition encoder $\tau$ that $c=\tau(I_c)$, is incorporated into the Diffusion UNet through a cross-attention mechanism.  The objective function of the UNet is defined as:
\begin{equation}
\mathcal{L}_{\text{SD}} = ||\epsilon - \epsilon_{\theta} (z_t, z_{lc}, m, c, t)||_2^2
\end{equation}

The condition $c$ is critical in the SD framework for inpainting tasks, as it provides essential clothing information for the process.
Therefore, the selection and training of the condition encoder $\tau$ are of utmost importance.
The Paint-by-Example approach \cite{Yang_Gu_Zhang_Zhang_Chen_Sun_Chen_Wen_2022} suggests using the CLIP \cite{Radford_Kim_Hallacy_Ramesh_Goh_Agarwal_Sastry_Amanda_Mishkin_Clark_et} image ViT encoder to extract this condition. However, our findings indicate that this is not the most effective strategy, given that the CLIP encoder is pre-trained on open-world images rather than clothing-specific images. Consequently, we propose fine-tuning the ViT encoder $\tau$ to yield a more accurate condition, tailored for clothing images.

\subsection{Finetuning ViT on Clothes}
Paint by Example utilizes CLIP to enhance detailed information, yet CLIP is inclined towards classification, leading to significant inter-class differences but minimal intra-class variation. This results in a lack of distinctiveness for items within the same category. Consequently, extracting effective information from clothing datasets becomes crucial for producing optimal conditions for SD to generate superior images. However, this is challenging due to the scarcity of annotations or other supervised methods beneficial for virtual try-on in clothing datasets.
Inspired by \cite{Caron_Touvron_Misra_Jegou_Mairal_Bojanowski_Joulin_2021}, which utilizes self-supervised learning for visual representation, we implement a similar approach. The core of this training methodology lies in contrasting representations from different perspectives of the same data instance. 

This concept is operationalized through a dual training setup comprising teacher and student networks. The teacher network $\tau$ produces pseudo-labels for the student network $\tau'$ by analyzing representations from augmented views of the input data. Notably, $\tau$ is maintained as the moving average of $\tau'$, ensuring more stable and consistent feature extraction. In parallel, the student network $\tau'$ is trained to conform to these pseudo-labels, striving to minimize similarity across representations from various instances.
This strategy facilitates the extraction of semantically rich features from clothing data, significantly enhancing the conditioning quality for SD.
The loss function of the finetuning process can be expressed as follows:
\begin{equation}
\mathcal{L}_{\text{SS}} = \sum_{i=1}^{M} \sum_{\substack{j=1 \ j \neq i}}^{N+M} H(\tau(I_{c_i}) , \tau'(I_{c_j})),
\end{equation}
where $M$ and $N$ represent the total number of global and local crops, respectively. The functions $\tau(\cdot)$ and $\tau'(\cdot)$ denote the ViT feature extraction process conducted by the teacher and student networks, respectively. The term $H(a, b) = -a \log b$ signifies information entropy. $I_{c_j}$ refers to the augmented view of the input data instance $I_c$.
The augmentation process begins with a random resized crop operation, applying a scale factor and bicubic interpolation. This is followed by flip and color jitter transformations. Additionally, the process incorporates Gaussian blur and normalization for enhanced diversity in the data representation. 
The scales for the local crops are specified in the range $[0.05, 0.25]$ of the whole image, while those for the global crops fall within the range $[0.25,1]$.

\begin{figure}[t]
\centering
\includegraphics[width=0.9\linewidth]{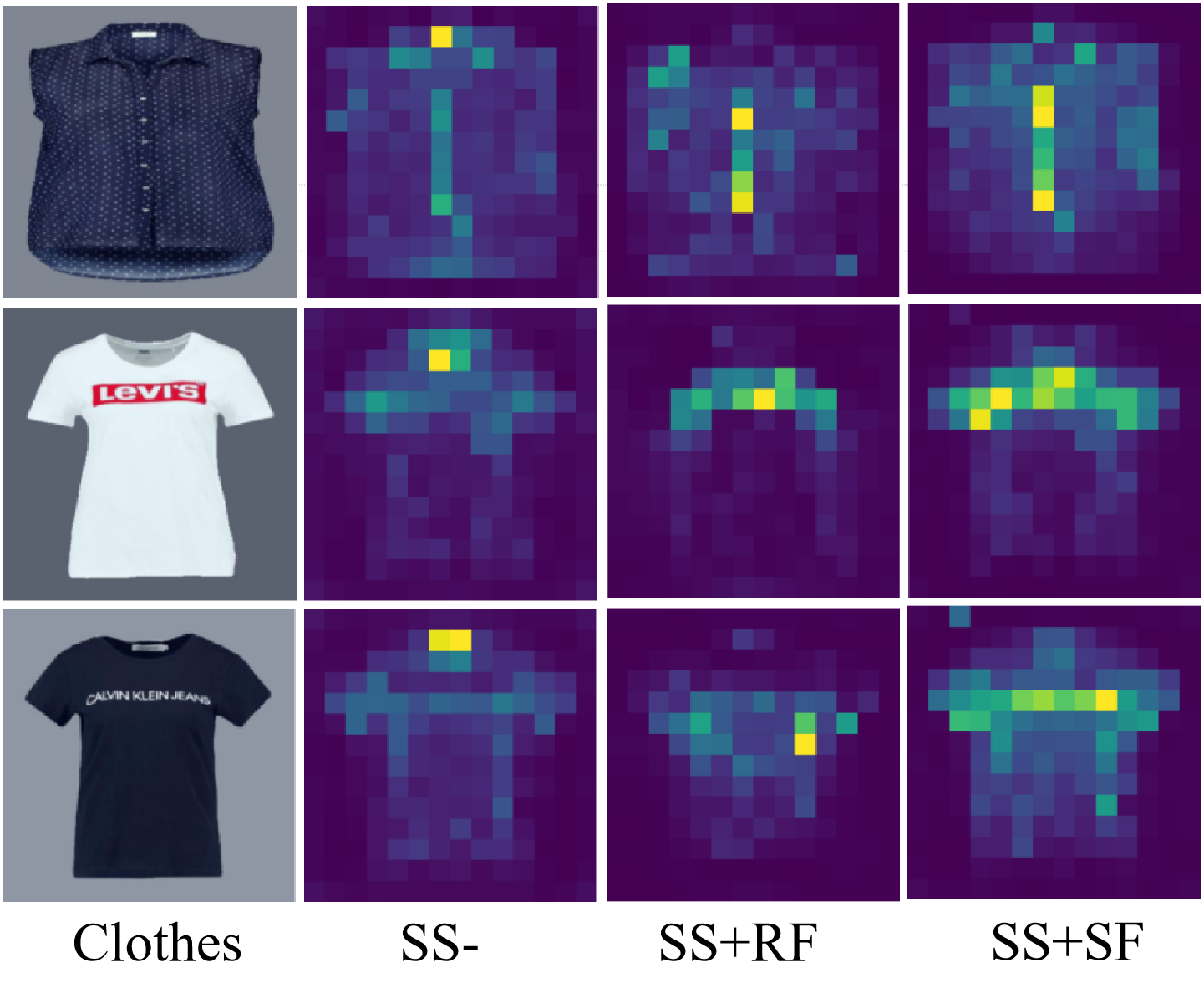}
\caption{Visualization of the Average Head's Attention for the Class Token in ViT. ``SS-'' represents the scenario without any finetuning, ``SS+RF'' indicates the use of random local crops for self-supervised finetuning, and 'SS+SF' signifies the application of our method, which involves selectively choosing local crops for self-supervised finetuning.}
\label{fig:attn}
\vspace{-16pt}
\end{figure}

The teacher network is utilized as the final condition encoder in our framework. In Figure~\ref{fig:attn}, we present visualizations of the average head attention corresponding to the class token in ViT, both before and after self-supervised fine-tuning. In these visualizations, ``SS-'' represents the state without fine-tuning, while ``SS+RF'' indicates the application of self-supervised fine-tuning. A comparative analysis between ``SS-'' and ``SS+RF'' demonstrates that self-supervised fine-tuning on our dataset results in heightened attention to specific key areas of the clothing.
Overall, this self-supervised approach empowers ViT to focus distinctively on each feature of the clothing. After undergoing self-supervised fine-tuning, ViT shows increased attention to particular local areas within the clothing images, reflecting an enhanced understanding of clothing features and details.

\begin{figure*}[t]
\centering
\includegraphics[width=0.95\linewidth]{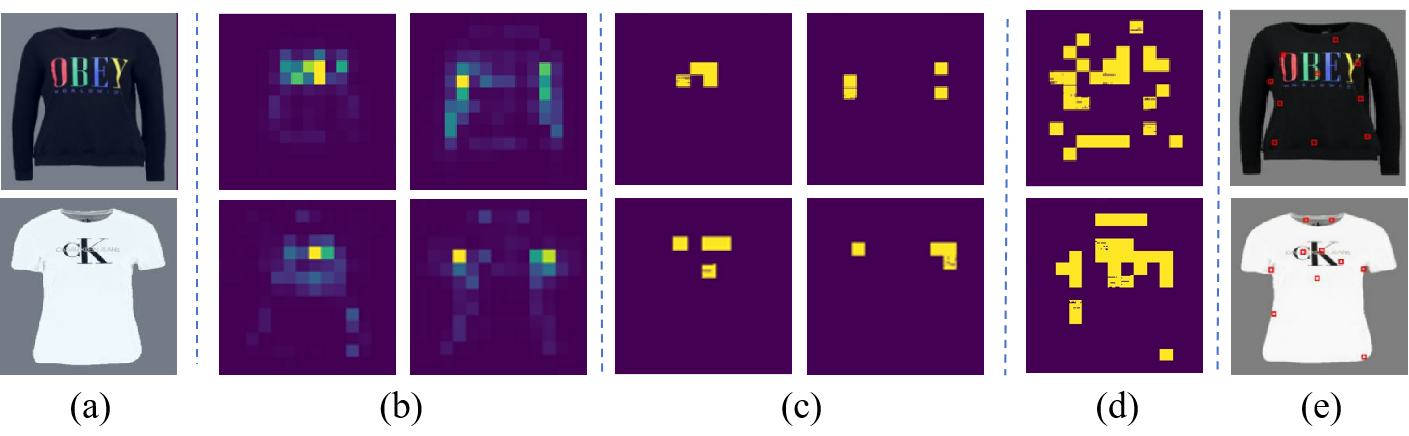}
\caption{In this visualization, (a) displays the original image input to the condition encoder $\tau$. Subfigure (b) illustrates the attention maps of two specific heads within the self-attention mechanism of ViT, highlighting areas of focus. Subfigure (c) shows the focal points derived from the attention maps presented in (b), pinpointing the specific areas receiving the highest attention. The aggregation of focal points across all heads is depicted in (d), demonstrating the comprehensive attention landscape. Based on the focal points in (d), clustering is conducted to identify key cluster centers, which are prominently marked in red in subfigure (e), indicating areas of significant attention across all heads.}
\label{fig:dino}
\vspace{-16pt}
\end{figure*}

\subsection{Finetuning ViT with Keypoints}
While we implement a self-supervised approach to ViT, we find that using a pre-trained ViT directly offers limited adaptability to our dataset and inadequate capability to capture clothing-specific features. 
During the initial stages of the fine-tuning process, local crops are randomly sampled. However, this approach often overlooks vital keypoints of the clothing, resulting in the omission of important details.
Given our objective for ViT to accurately identify key elements of clothing, such as collars, sleeves, text, and patterns, we intentionally fine-tune ViT on our dataset to enhance its attentiveness to these critical aspects. 
To overcome the shortcomings of random sampling, we refine our methodology to selectively sample local crops that encompass these essential keypoints. This adjustment ensures that the network is furnished with the detailed information, enabling a more precise representation of clothing.

In the self-attention maps of the ViT, we observe that different heads focus on distinct segments of clothing, closely aligning with key areas. Figure~\ref{fig:dino} showcases this, where subfigures (b) and (c) accurately target text and sleeves on the clothing. Consequently, we compile all points with attention surpassing a specified threshold, referred to as high-attention points, from each head's attention and merge them into a singular map, as depicted in subfigure (d). We then cluster these points based on their proximity, ensuring the number of clusters corresponds to the quantity of local crops. As revealed in Figure~\ref{fig:dino} (e), the centroids of these clusters effectively encompass vital garment areas like sleeves, collars, hems, and patterns.
These identified centroid points are then utilized as the basis for generating new local crops. Following this, we apply a range of data augmentation techniques to derive the final local crops, which are integral to the fine-tuning of ViT on clothes dataset. After fine-tuning, the stabilized ViT model is incorporated into the training of SD inpainting network, enhancing its effectiveness.

Specifically, we identify 10 centroids in the clustering process as the keypoints for each clothing item. The local crops are then centered around these keypoints, maintaining a $[0.05, 0.25]$ ratio to the entire image. 
In Figure~\ref{fig:attn}, 'SS+SF' denotes the application of our method in selecting local crops for self-supervised fine-tuning. Comparing ``SS-'' and ``SS+RF'', it is apparent that our dataset's self-supervised fine-tuning leads to increased attention on specific key areas. However, 'SS+SF' demonstrates a more pronounced focus on crucial points, text, patterns, and other intricate details on the clothing. This observation confirms that our self-supervised approach enables ViT to more effectively extract and emphasize the unique characteristics of clothing compared to previous methods.

\section{Experiments}

\subsection{Datasets}
Our experimental evaluations primarily utilize the VITON-HD dataset \cite{Choi_Park_Lee_Choo_2021}, comprising 13,679 pairs of frontal-view women and top clothing images. Following precedent \cite{Choi_Park_Lee_Choo_2021,Lee_Gu_Park_Choi_Choo_2022}, we partition the dataset into training and test sets, encompassing 11,647 and 2,032 pairs, respectively. Experiments are conducted at $512\times384$. Furthermore, to validate the robustness of our method in more intricate scenarios, we extend our experiments to include the the DressCode dataset \cite{Morelli_Fincato_Cornia_Landi_Cesari_Cucchiara}. Detailed results of these extended evaluations will be provided in the supplementary material.

\subsection{Evaluation Metrics}
To evaluate the effectiveness of our method in different test scenarios, we employ a range of metrics. In the paired setting, where a clothing image is used to reconstruct a person's image, we use two widely accepted metrics: Structural Similarity (SSIM) \cite{Wang_Bovik_Sheikh_Simoncelli_2004} and Learned Perceptual Image Patch Similarity (LPIPS) \cite{Zhang_Isola_Efros_Shechtman_Wang_2018}. 
For the unpaired setting, involving the alteration of the person's clothing, we measure the performance using Frechet Inception Distance FID \cite{Heusel_Ramsauer_Unterthiner_Nessler_Hochreiter_2017} and Kernel Inception Distance KID \cite{kid}.
To capture the complex aspects of human perception, we have incorporated a user study into our evaluation process. Specifically, we have created composite images using various methods for 200 randomly selected pairs from our test set, each at a resolution of $512\times384$. These images were then evaluated by a panel of 30 human judges who were tasked with identifying the method that most effectively restored clothing details and produced the most lifelike results for each pair. The frequency of each method being chosen as the best performer in these two aspects is thoroughly documented, providing a detailed insight into our method's performance.

\subsection{Implementation Details}

The training of our network is systematically divided into two distinct stages. In the initial stage, we leverage local crops derived from attention maps to train the ViT model. We use 10 local crops and opt for the VIT-base model with a patch size of 16. The training commences with an initial learning rate of 2e-5, and a batch size is set at 32. This model undergoes training for more than 30 epochs on our dataset.
In the second stage, we employ the pre-trained ViT model with its parameters fixed and focus on training the entire network. For this phase, the learning rate is adjusted to 1e-5, and the batch size is reduced to 8. This stage of training extends over 40 epochs, allowing for comprehensive learning and adaptation. Overall, the training process requires only 6 hours to finetune ViT and 24 hours to finetune SD, which is significantly faster compared to full-scale SD training.
Besides, the testing is conducted on a single 3090 GPU. And we adopt PLMS Sampling for inference with 100 steps. 
\textit{It is crucial to highlight that our model does not incur any additional inference time to SD. The average generation time per image is 6.31 seconds, which is roughly equivalent to that of PbE with 6.28 seconds.}

\begin{figure*}[t]
\centering
\includegraphics[width=0.94\linewidth]{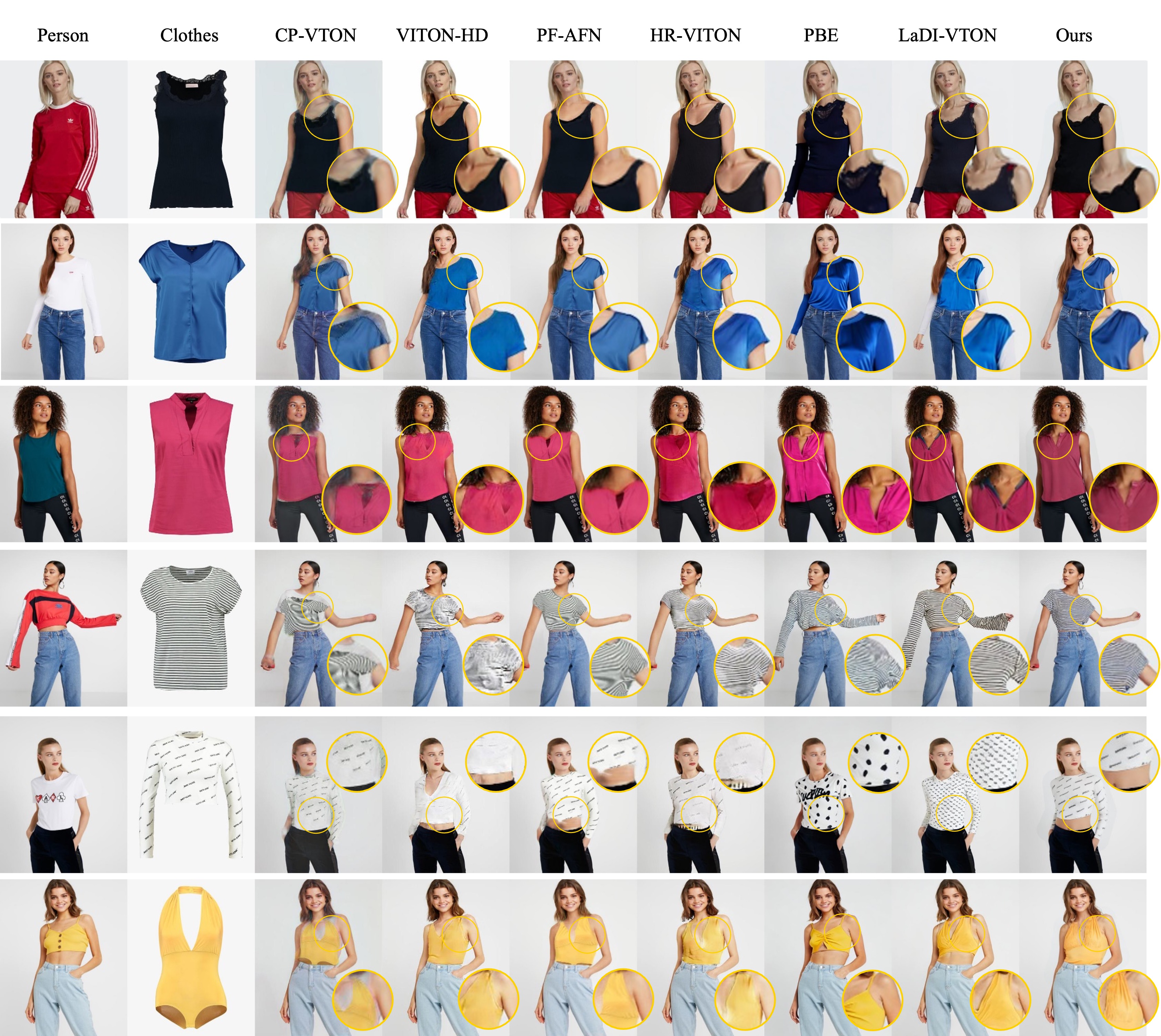}
\caption{Qualitative comparisons.}
\label{fig:fine}
\end{figure*}

\begin{table}
\centering
\setlength{\tabcolsep}{3pt}
\begin{tabular}{lccccc}
\toprule
Method           & SSIM↑  & LPIPS↓ & FID↓   & KID↓  & User↑  \\ \midrule
CP-VTON          & 0.791 & 0.141 & 30.25 & 4.012 &  0.006/0.003 \\
VITON-HD         & 0.843 & 0.076 & 11.64 & 0.300 &   0.061/0.058\\
PF-AFN           & 0.858 & 0.082 & 11.30 & 0.283 &   0.273/0.187\\
HR-VITON         & 0.878 & 0.061 & 9.90  & 0.188 &   0.121/0.071\\
PbE & 0.843 & 0.087 & 10.15 & 0.204 &   0.003/0.038\\
LaDI-VTON        & 0.876 & 0.059 & 9.07  & 0.148 &   0.067/0.121\\ \midrule
Ours             & \textbf{0.886} & \textbf{0.052} & \textbf{8.93}  & \textbf{0.117} &    \textbf{0.469}/\textbf{0.522}\\ \bottomrule
\end{tabular}
\caption{Quantitative comparisons. KID has been multiplied by 100 to facilitate comparison. The user study scores indicate the frequencies of method selection based on two distinct evaluation criteria: similarity of reconstruction (left) and realism (right).}
\label{tab:baseline}
\vspace{-16pt}
\end{table}

\subsection{Quantitative Evaluations}
Our target baselines are CPVTON \cite{Wang_Zheng_Liang_Chen_Lin_Yang_2018}, PF-AFN \cite{Ge_Song_Zhang_Ge_Liu_Luo_2021}, VITON-HD \cite{Choi_Park_Lee_Choo_2021} and HR-VTON \cite{Lee_Gu_Park_Choi_Choo_2022}, and diffusion methods Paint-by-Example (PbE) \cite{Yang_Gu_Zhang_Zhang_Chen_Sun_Chen_Wen_2022} and LaDI-VTON \cite{morelli2023ladi}.
The comparison results on VITON-HD dataset are shown in Table~\ref{tab:baseline}. 
The performance of traditional methods such as CP-VTON, VITON-HD and PF-AFN is relatively inferior, while the PbE method demonstrates acceptable results after finetuning on the dataset. Furthermore, the HR-VITON method, which performs warp operations on high-resolution images, exhibits favorable outcomes. Similarly, LaDI-VTON, which utilizes textural inversion to map images to image clip embeddings, also yields promising results.
While our method achieves superior performance, attributed to the effective utilization of self-supervised learning techniques, which enables the network to focus on details and key points, thereby addressing the shortcomings of previous models.

We further compare our method against competitive approaches, namely  PF-AFN~\cite{Ge_Song_Zhang_Ge_Liu_Luo_2021}, HR-VTON~\cite{Lee_Gu_Park_Choi_Choo_2022}, and LaDI-VTON~\cite{morelli2023ladi} on the DressCode dataset~\cite{Morelli_Fincato_Cornia_Landi_Cesari_Cucchiara}. Table~\ref{tab:dress} shows that our method achieves the best results across different categories, including upper, lower, and dresses, in terms of similarity and realism metrics. These qualitative results demonstrate the effectiveness of our method on a diverse range of datasets, highlighting its strong transferability and ability to adapt to various clothing categories.

\subsection{Qualitative Evaluations}
In Figure \ref{fig:fine}, we present qualitative comparisons of our proposed method with other state-of-the-art methods on VITON-HD dataset.
The analysis reveals distinct differences in performance. CP-VTON, for instance, tends to produce results that appear more artificial and less authentic in terms of integration. VTON-HD, on the other hand, falls short in capturing textural details. Compared to these methods, PF-AFN and HR-VTON come closer to replicating the original images but still exhibit deficiencies in several detail aspects. Paint-by-Example (PbE) shows some ability to learn similar features but struggles with capturing variations between individuals. LaDI-VTON, while showing promise, faces challenges similar to PbE.

\begin{figure*}[t]
\centering
\includegraphics[width=0.9\linewidth]{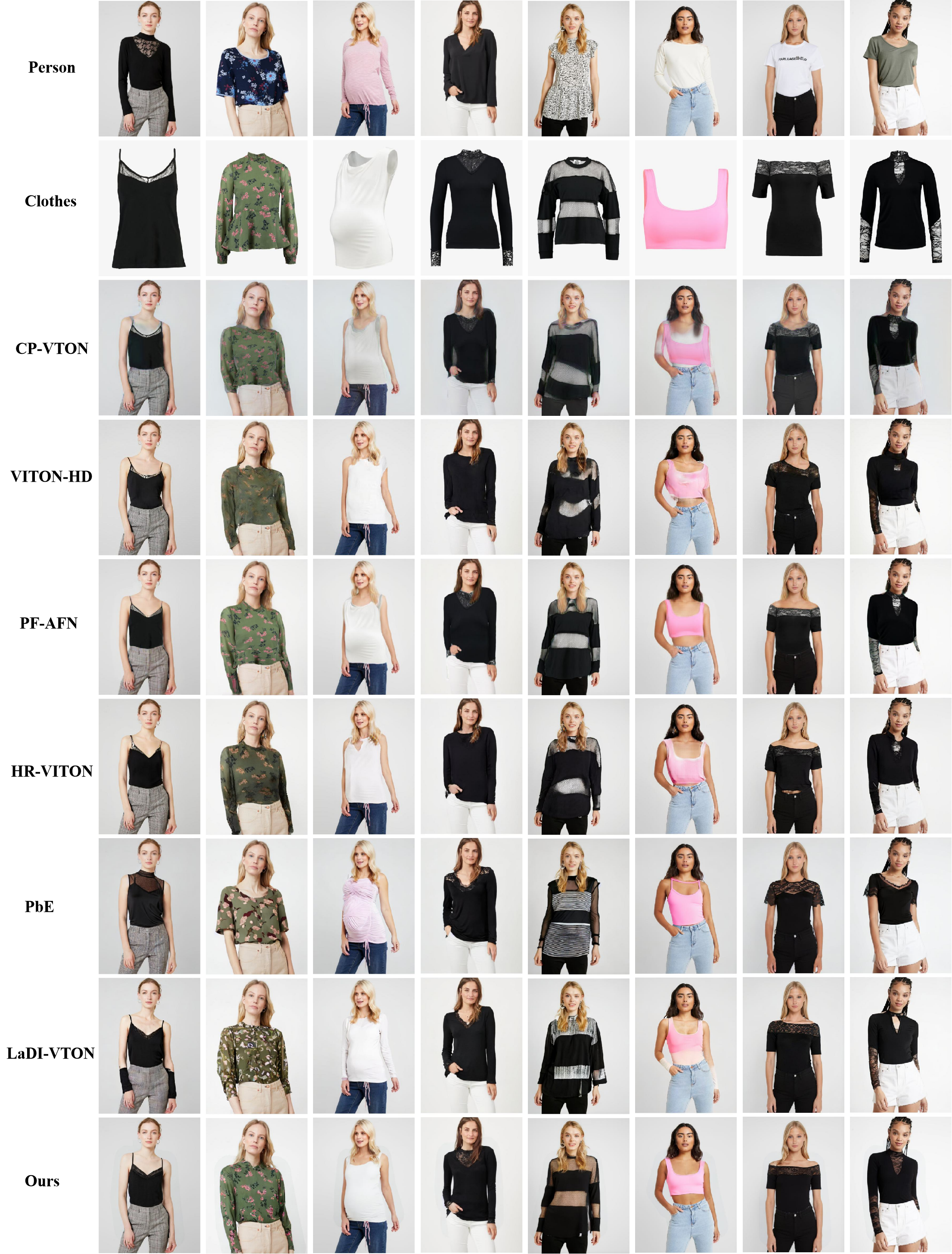}
\caption{More Qualitative Comparisons with Baselines on VITON-HD Dataset.}
\label{fig:sup_base}
\end{figure*}

Our method stands out in handling numerous details more effectively. Particular attention has been given to key areas such as collars and hems, as evidenced in the first and third rows of the figure. Additionally, our approach more accurately reflects the overall texture and material of the original images, as seen in the second and fourth rows, and demonstrates a slight superiority in handling various intricate details, as shown in the fifth and sixth rows. These successes are largely due to our fine-tuning process, which has enhanced the network's ability to concentrate on crucial details at different positions, thereby improving overall precision. More visual results are shown in figure~\ref{fig:sup_base}.

\begin{figure*}[t]
\centering
\includegraphics[width=0.95\linewidth]{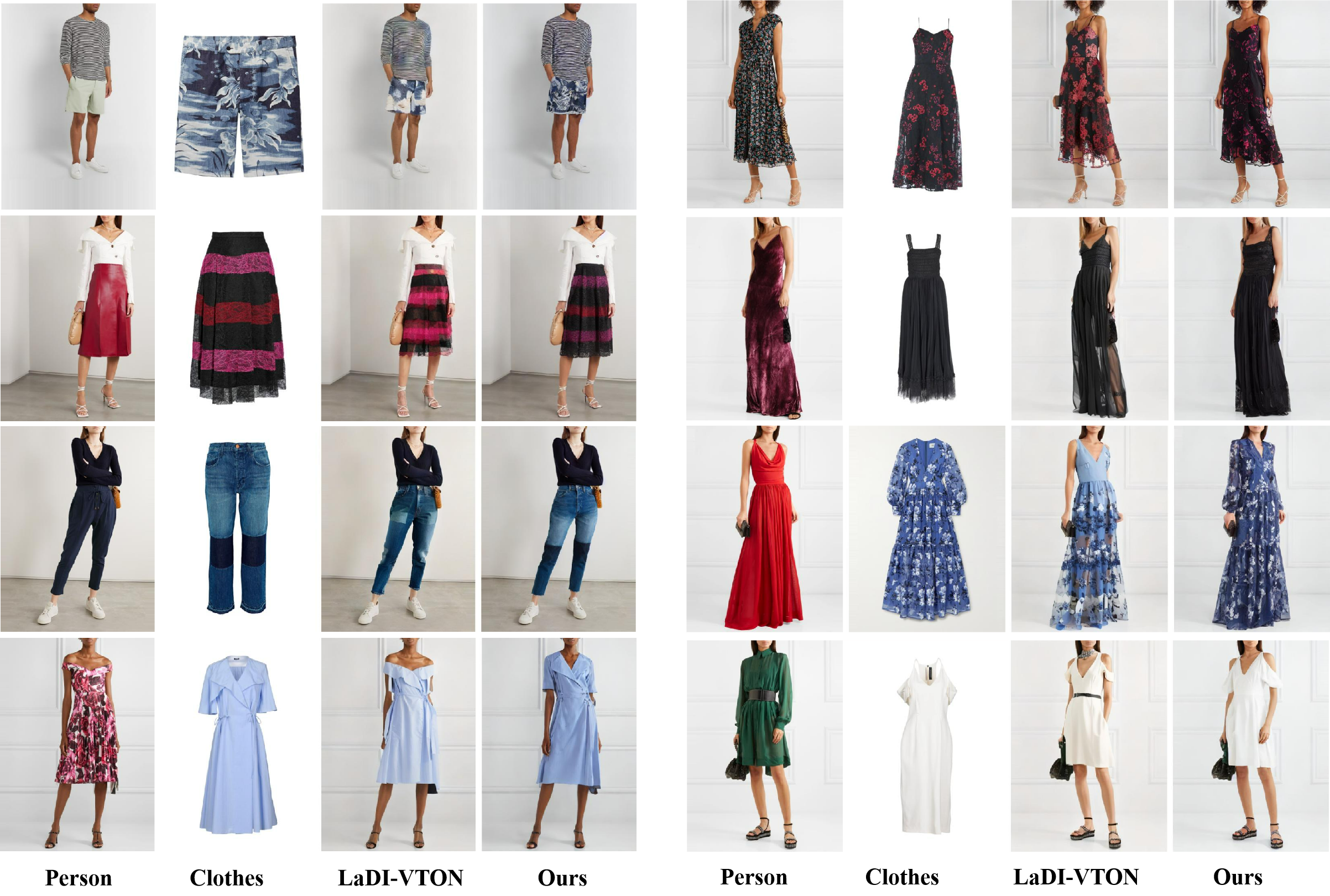}
\caption{Qualitative Comparisons with Competitive Baseline on DressCode Dataset.}
\label{fig:sup_dress}
\end{figure*}

In order to compare the visual results of our method on DressCode Dataset with the highest-scoring baseline LaDI-VTON~\cite{morelli2023ladi} in terms of evaluation metrics, we present a visual comparison in Figure\ref{fig:sup_dress}. This comparison allows for a direct assessment of the performance of our method against the state-of-the-art baseline. 
The results obtained from our proposed method represent a significant breakthrough in the field of clothing generation. Specifically, we have demonstrated the ability to achieve an exceptional level of fidelity, encompassing various intricate texture patterns and other essential features, that faithfully replicates the target garments. Notably, our approach also exhibits a heightened realism in terms of the overall fit and texture of the generated clothing, which further accentuates the superiority of our method over existing approaches.

\begin{table*}[]
\centering
\begin{tabular}{c|cc|cc|cc}
\hline
\multirow{2}*{Method} & \multicolumn{2}{c|}{Upper} & \multicolumn{2}{c|}{Lower} & \multicolumn{2}{c}{Dresses} \\ \cline{2-7} 
                        & LPIPS↓        & FID↓         & LPIPS↓        & FID↓         & LPIPS↓         & FID↓         \\ \hline
PF-AFN                  & 0.0380       & 14.32       & 0.0445       & 18.32       & 0.0758        & 13.59       \\
HR-VITON                & 0.0635       & 16.86       & 0.0811       & 22.81       & 0.1132        & 16.12       \\
LaDI-VTON               & 0.0249       & 13.26       & 0.0317       & 14.79       & 0.0442        & 13.40       \\
Ours                    &    0.0216          &   10.65          & 0.0255       & 11.89       &      0.0382         &  10.76           \\ \hline
\end{tabular}
\caption{Quantitative comparisons with competitive baselines on DressCode dataset.}
\label{tab:dress}
\vspace{-16pt}
\end{table*}

\begin{table}
\centering
\begin{tabular}{lcccc}
\toprule
Method           & SSIM↑  & LPIPS↓   \\\midrule
(a) CLIP Encoder   & 0.841 & 0.142 \\
(b) SS- Encoder  & 0.847 & 0.107 \\
(c) SS+RF Encoder   & 0.860 & 0.105  \\
(d) SS+SF Encoder     & 0.862 & 0.104 \\
(e) Final Network              & 0.886 & 0.052  \\ \bottomrule
\end{tabular}
\caption{Results of Ablation Studies}
\label{tab:abla}
\vspace{-16pt}
\end{table}

\subsection{Ablation Studies}

In our ablation studies, we assess the impact of each individual enhancement on the overall performance. These experiments utilize images of size $512\times384$. The CLIP encoder has a patch size of 14 and an image embedding size of 1024, while ViT has a patch size of 16 and an image embedding of 768.  To ensure fairness, we maintain consistency by mapping the embedding of ViT to 1024 dimensions using a linear layer. We compare the following configurations: (a) CLIP: Utilizing CLIP as the encoder for conditioning. (b) SS-Encoder: Employing ViT for conditioning through self-supervised learning, without any fine-tuning. (c) SS+RF Encoder: Conditioning with ViT post fine-tuning using random local crops. (d) SS+SF Encoder: Conditioning with ViT after fine-tuning with selected local crops based on self-attention maps. Lastly, (e) the Final Network: Our complete network, integrating warping to interact with the features.

As shown in Table~\ref{tab:abla}, it is evident that ViT, when subjected to self-supervised training, demonstrates notably superior improvement in comparison to CLIP. Notably, the application of fine-tuning with random local crop on our dataset fails to yield substantial improvements. 
However, our proposed fine-tuning method encompasses pivotal considerations, facilitating a more comprehensive capture of intricate clothing features and consequently leading to further enhancements in the experimental outcomes. 
Lastly, our comprehensive network analysis reveals that the incorporation of warping for interactions with ViT features highlights an inadequate focus of UNet features on ViT.
Subsequent to the introduction of warp features, UNet features exhibit a notable improvement in effectively attending to the corresponding ViT feature regions, thereby significantly elevating the overall quality of the results.

\begin{figure*}[t]
\centering
\includegraphics[width=0.9\linewidth]{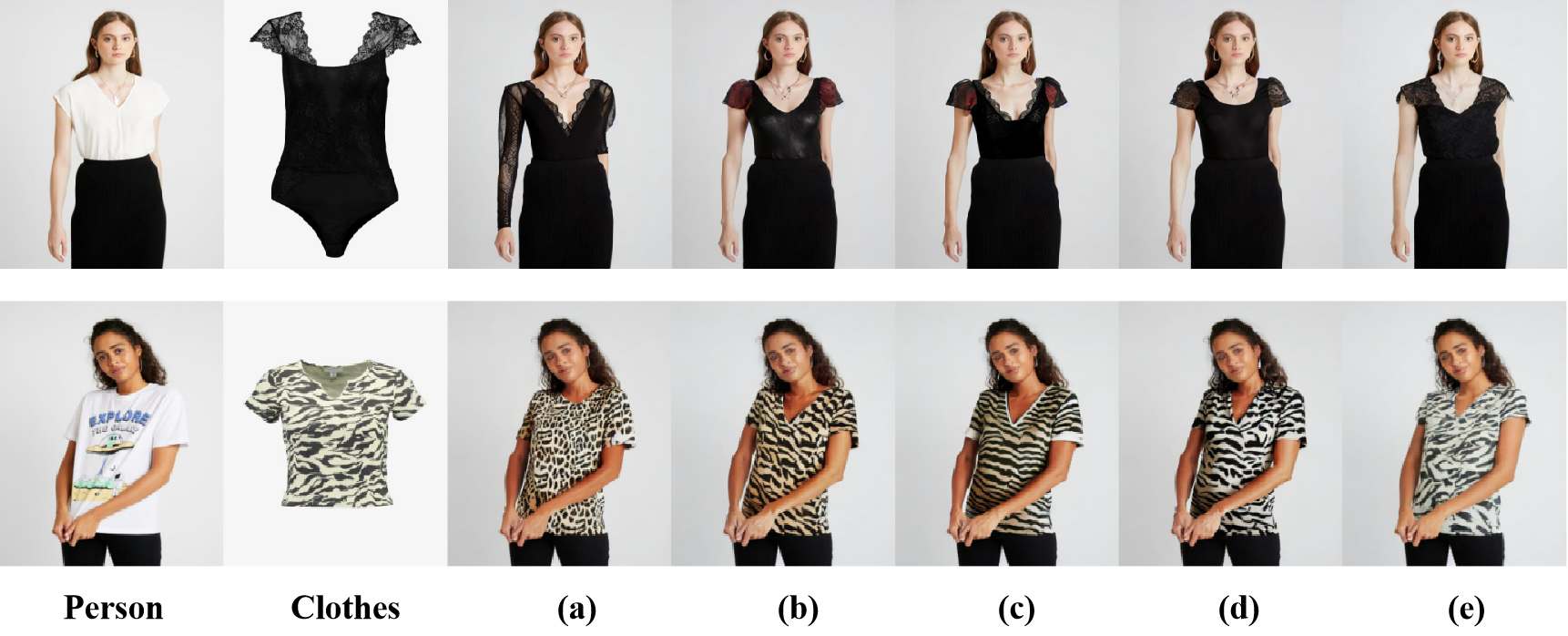}
\caption{Visualization of ablation studies.}
\label{fig:sup_abl}
\end{figure*}

As shown in Figure~\ref{fig:sup_abl}, the results clearly indicate that ViT, when subjected to self-supervised training, exhibits a significantly superior improvement compared to CLIP. It is worth noting that 
our proposed fine-tuning method takes critical factors into account, allowing for a more comprehensive capture of intricate clothing features, consequently leading to further enhancements in the experimental outcomes. Moreover, our comprehensive network analysis reveals that the incorporation of warping for interactions with ViT features highlights an insufficient focus on UNet features on ViT. Following the introduction of warp features, UNet features demonstrate a noticeable improvement in effectively attending to the corresponding ViT feature regions, thereby significantly elevating the overall quality of the results.

\section{Limitations}
\begin{figure}[t]
\centering
\includegraphics[width=0.9\linewidth]{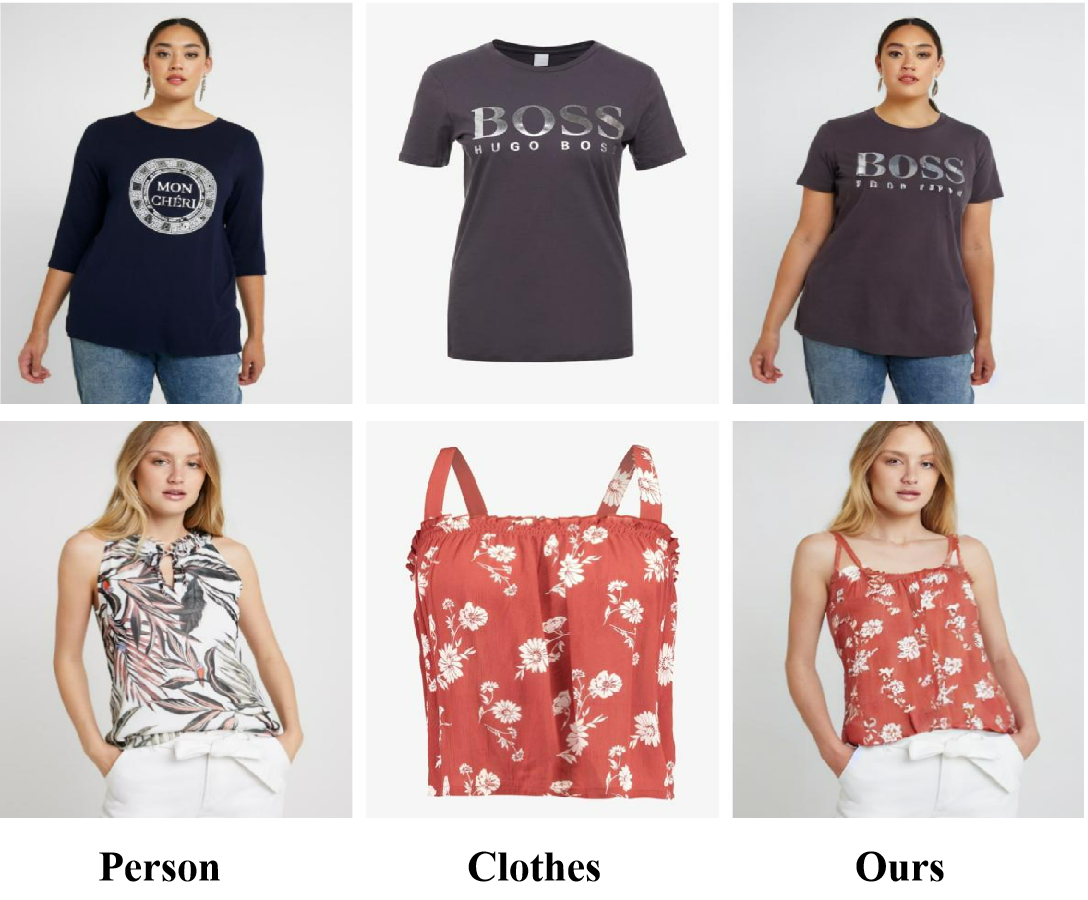}
\caption{Visualization of Limitations of our method.}
\label{fig:limit}
\vspace{-16pt}
\end{figure}

The results presented in Figure~\ref{fig:limit} highlight certain unresolved issues with our proposed approach. Specifically, we have observed that the effectiveness of our method is not yet perfect when it comes to addressing minor details (as evidenced by the first row). This limitation is not unique to our approach, as it represents a common bottleneck for most current methods in the field. In addition, we have also noted that providing two-dimensional guidance for laying out clothes may potentially mislead the three-dimensional construction process, as illustrated by the shoulder straps in the second row. While these challenges remain, we are committed to exploring potential solutions to address them in future works.

\section{Conclusions}
In this paper, we present an innovative and effective methodology for virtual clothes try-on. The method integrates a self-supervised ViT with a diffusion model.  It focuses on enhancing details by comparing local and global clothing image embeddings from ViT, demonstrating an acute understanding of complex visual elements. Techniques like conditional guidance, focus on key regions, and specialized content loss contribute to its thoroughness. These strategies enable the diffusion model to accurately replicate clothing details, significantly enhancing the realism and clarity of virtual try-on experiences.

{\small
\bibliographystyle{ieee_fullname}
\bibliography{egbib}
}

\end{document}